\documentclass[11pt,a4paper]{article}

\usepackage[utf8]{inputenc}
\usepackage[T1]{fontenc}
\usepackage{lmodern}
\usepackage[margin=2.5cm]{geometry}
\usepackage{microtype}
\usepackage{amsmath,amssymb}
\usepackage{graphicx}
\usepackage{booktabs}
\usepackage{array}
\usepackage{siunitx}
\usepackage[font=small,labelfont=bf]{caption}
\usepackage{enumitem}
\usepackage[numbers,sort&compress]{natbib}
\usepackage{xcolor}
\usepackage[colorlinks=true,linkcolor=blue!50!black,citecolor=blue!50!black,urlcolor=blue!50!black]{hyperref}

\graphicspath{{figures/}}
\newcommand{\dlo}{DLO-Anything}

\title{\textbf{DLO-Anything: Towards a Foundation Model for\\ Deformable Linear Objects}}

\author{%
  Ludwig Streloke\textsuperscript{a}\thanks{Corresponding author: \texttt{ludwig.streloke@faps.fau.de}},
  Matthias Lang\textsuperscript{a},
  Christopher Straub\textsuperscript{b},\\
  Lukas Zeh\textsuperscript{c},
  Alexander Weilacher\textsuperscript{a},
  Annalena Hartmann\textsuperscript{a},
  J\"org Franke\textsuperscript{a}\\[10pt]
  {\fontsize{9.5}{12}\selectfont
  \textsuperscript{a}Friedrich-Alexander-Universit\"at Erlangen-N\"urnberg,}\\
  {\fontsize{9.5}{12}\selectfont
  Institute for Factory Automation and Production Systems (FAPS),}\\
  {\fontsize{9.5}{12}\selectfont
  Egerlandstra{\ss}e 7--9, 91058 Erlangen, Germany}\\[6pt]
  {\fontsize{9.5}{12}\selectfont
  \textsuperscript{b}Fraunhofer Institute for Integrated Systems}\\
  {\fontsize{9.5}{12}\selectfont
  and Device Technology (IISB),}\\
  {\fontsize{9.5}{12}\selectfont
  Schottkystra{\ss}e 10, 91058 Erlangen, Germany}\\[6pt]
  {\fontsize{9.5}{12}\selectfont
  \textsuperscript{c}University of Stuttgart,}\\
  {\fontsize{9.5}{12}\selectfont
  Institute for Control Engineering of Machine Tools and Manufacturing Units (ISW),}\\
  {\fontsize{9.5}{12}\selectfont
  Seidenstra{\ss}e 34, 70174 Stuttgart, Germany}
}

\date{}

\begin{document}
\maketitle

\begin{abstract}
\noindent
Predicting the dynamics of deformable linear objects (DLOs) remains challenging due to varying physical properties and complex contact interactions.
Existing physics-based and learned models often require object-specific calibration or adaptation, limiting their transferability.
This paper introduces \dlo{}, an approach toward a deep learning-based foundation model for DLO dynamics.
A physics-conditioned transformer integrates DLO states, material properties, robot actions, and potential rigid and soft contacts through a unified token-based representation, supporting dynamics prediction and inverse parameter identification.
A scalable simulation-based training pipeline generates diverse manipulation scenarios for model pretraining.

An empirical scaling study examines generalization to unseen robot motions as a function of training data volume and motion diversity.
The results reveal a diversity-dependent error floor: increasing the number of motion categories from one to eight reduces the prediction error over six-second autoregressive rollouts by \SI{30}{\percent}, while additional scenes of the same categories yield diminishing improvements.
A joint empirical scaling law captures both effects, providing initial evidence for scalable, physics-conditioned DLO dynamics modeling.
\end{abstract}

\begin{figure}[t]
    \centering
    \includegraphics[width=\linewidth]{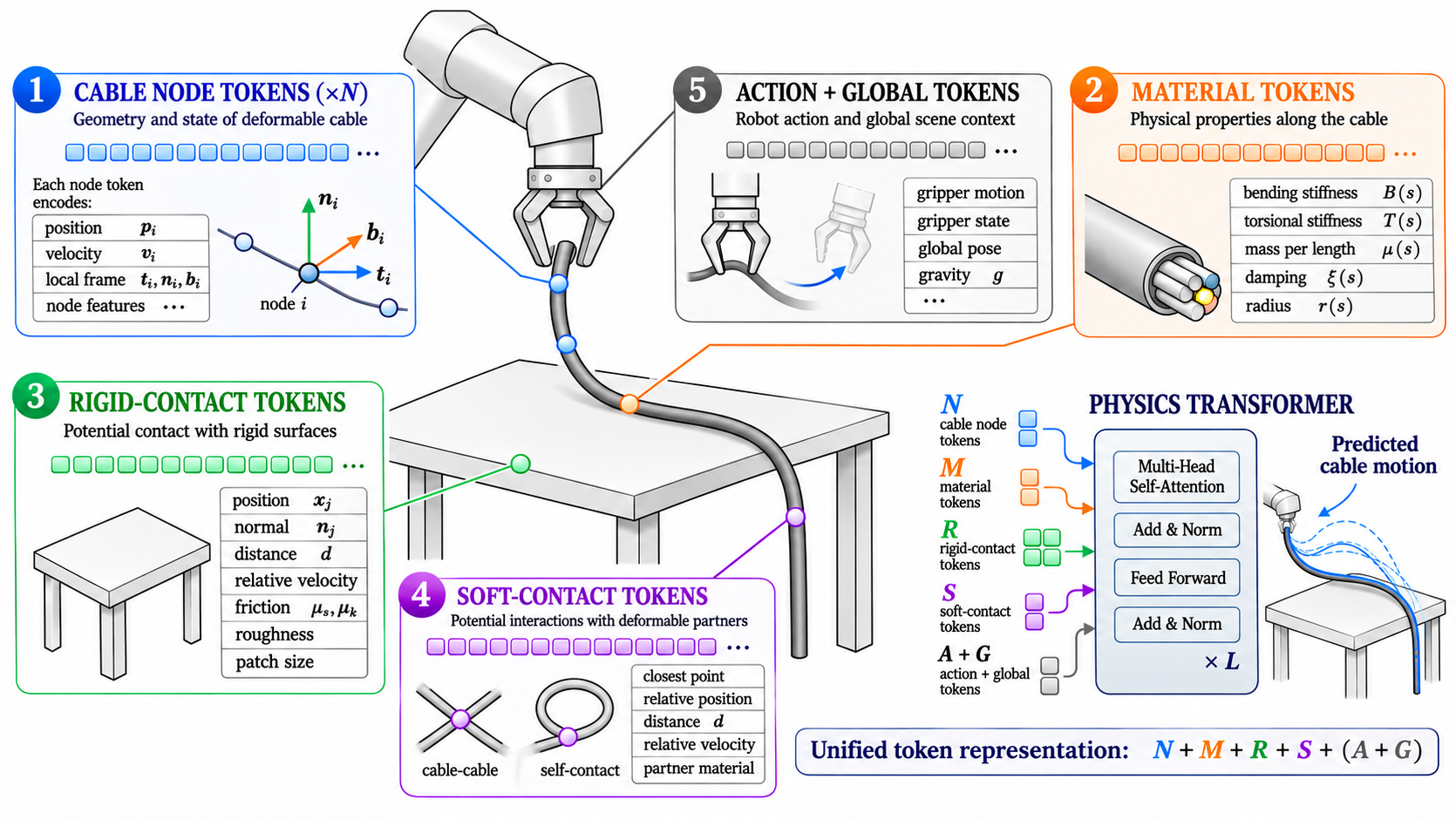}
    
    \caption{The \dlo{} concept. A physics-conditioned transformer integrates cable states, material properties, planned robot actions, and potential rigid and soft contacts through a unified token-based representation to predict DLO dynamics.}

    \label{fig:concept}
\end{figure}

\section{Introduction}
\label{sec:introduction}

Accurately predicting the behavior of deformable linear objects (DLOs), such as cables, wires, and hoses, remains a major challenge in robotic manipulation~\cite{caporali2026survey}.
Unlike rigid objects, DLOs exhibit complex dynamics governed by their geometry, material properties, and interactions with the environment.
During manipulation, cables may slide across surfaces, bend around obstacles, or come into contact with themselves and other objects.
These interactions introduce friction and changing contact forces, making accurate predictions particularly challenging.

Existing approaches rely on physics-based simulators, such as Discrete Elastic Rods and Cosserat rod models~\cite{bergou2008der,gazzola2018cosserat}, or learned dynamics models~\cite{yang2022differentiable,yu2023global}.
While physics-based approaches are computationally expensive and require careful parameter calibration, learned models frequently depend on object-specific training.
Recent parameter-conditioned approaches improve transferability~\cite{caporali2024online,zhang2024adaptigraph}, but generalization across different cable properties and contact conditions remains challenging.

Foundation models offer a promising direction by learning transferable representations from diverse data~\cite{bommasani2021foundation}.
Models such as Segment Anything~\cite{kirillov2023sam} and Depth Anything~\cite{yang2024depthanything} have demonstrated broad generalization in computer vision.
In scientific machine learning, approaches such as Multiple Physics Pretraining~\cite{mccabe2024mpp}, Poseidon~\cite{herde2024poseidon}, and Walrus~\cite{mccabe2026walrus} extend the foundation model paradigm toward transferable prediction of physical dynamics.

Motivated by these developments, we introduce \dlo{} (Figure~\ref{fig:concept}), a novel approach toward a foundation model for contact-rich DLO dynamics.
The proposed architecture represents cable states, physical properties, robot actions, and environmental interactions as distinct tokens within a physics-conditioned transformer.
In particular, soft-contact tokens provide information about potential interaction partners, allowing the model to learn their influence on cable motion without requiring an explicit contact-force estimator.

\medskip
\noindent The main contributions of this paper are:
\begin{enumerate}[leftmargin=*,itemsep=4pt,topsep=5pt]
  \item \textbf{DLO-Anything concept:}
  A novel approach toward a foundation model for DLO dynamics that incorporates rigid and soft contact interactions and supports both forward prediction and inverse parameter identification.

  \item \textbf{Transformer-based architecture:}
  A physics-conditioned transformer that integrates cable states, physical properties, robot actions, and environmental interactions through a unified token-based representation, enabling the modeling of contact-rich cable dynamics.

  \item \textbf{Scalable training pipeline:}
  A simulation-based pipeline for generating diverse DLO dynamics data across material properties, geometries, manipulation actions, and contact scenarios to support large-scale model pretraining.

  \item \textbf{Scaling laws and generalization:}
  An empirical investigation of how training data volume and diversity affect prediction accuracy and generalization.
\end{enumerate}

\medskip
\noindent The remainder of this paper is organized as follows.
Section~\ref{sec:related} reviews related work.
Section~\ref{sec:concept} introduces the \dlo{} concept, followed by the transformer-based architecture in Section~\ref{sec:implementation}.
Section~\ref{sec:pipeline} describes the simulation-based data generation and training pipeline.
Section~\ref{sec:scaling} investigates how training data volume and robot motion diversity affect generalization to unseen manipulation trajectories.
Finally, Sections~\ref{sec:discussion} and~\ref{sec:conclusion} discuss the limitations, future research directions, and conclusions.

\section{Related work}
\label{sec:related}

\paragraph{Physics-based DLO models.}
Discrete Elastic Rods~\cite{bergou2008der} and Cosserat formulations~\cite{gazzola2018cosserat} are the standard descriptions of slender elastic bodies, and MuJoCo~\cite{mujoco} provides a cable model with bending and torsional stiffness that we use as the data source.
Chen et al.\ integrate Discrete Elastic Rods into MuJoCo and identify stiffness parameters of real cables~\cite{chen2025der}.
These models are interpretable and accurate after calibration, and the calibration effort per cable is their limit~\cite{streloke2027benchmark}.

\paragraph{Learned and parameter-conditioned dynamics.}
Graph networks learn particle and mesh dynamics from simulation~\cite{sanchezgonzalez2020gns,pfaff2021meshgraphnets}.
For DLOs, learned models have been used for shape control with offline training and online adaptation~\cite{yang2022differentiable,yu2023global}.
Closest to our proposal are models that take physical parameters as input: Caporali et al.\ condition a network on stiffness, damping and mass and estimate them online~\cite{caporali2024online}, AdaptiGraph conditions a graph network on a material variable found by few-shot optimisation~\cite{zhang2024adaptigraph}, and PhysWorld adapts the physical inputs of a learned world model to real videos~\cite{yang2025physworld}.
PhysTwin~\cite{jiang2025phystwin} and RopeDreamer~\cite{missal2026ropedreamer} learn action-conditioned predictors of single objects with implicit material.
\dlo{} belongs to the parameter-conditioned family and differs in scope: a broad parameter space instead of one object, a token interface that is meant to grow, and an explicit treatment of which parameters can be recovered.

\paragraph{Foundation models and scaling.}
Foundation models are trained on broad data and adapted to many tasks~\cite{bommasani2021foundation}; examples in vision are Segment Anything~\cite{kirillov2023sam} and Depth Anything~\cite{yang2024depthanything}, in robotics the cross-embodiment models of Open X-Embodiment~\cite{oxe2024}, and in physics multi-system pretraining~\cite{mccabe2024mpp,herde2024poseidon}.
Their progress is commonly described by scaling laws in data, parameters and compute~\cite{kaplan2020scaling,hoffmann2022chinchilla}; in robot imitation learning, diversity of the data has been found to matter more than its amount~\cite{lin2025datascaling}.
We report a first scaling study for DLO dynamics in Section~\ref{sec:scaling}.

\paragraph{Parameter identification and excitation.}
Identifying physical parameters from observed motion requires sufficiently informative excitation trajectories, a well-established challenge in robot dynamics~\cite{gautier1991,swevers1997}.
Recent approaches extend this principle to learned dynamics models~\cite{memmel2024asid}.
For deformable objects, Jakobsson et al.\ estimate rope parameters from a short excitation motion~\cite{jakobsson2026wiggle}, while Chen et al.\ infer external forces acting on a DLO from its observed shape~\cite{chen2026force}.
However, parameter identification is fundamentally limited by identifiability: parameters that produce indistinguishable observations cannot be uniquely recovered~\cite{bellman1970}.

\section{\dlo{} concept}
\label{sec:concept}

This section introduces the token-based representation of DLO dynamics, including physical properties, robot actions, and potential contact interactions. It further describes its use for forward dynamics prediction and inverse parameter identification.

\subsection{Token-based representation of DLO dynamics}

The proposed token representation provides a comprehensive structure for describing DLO dynamics across different physical properties, manipulation actions, and contact configurations.
Figure~\ref{fig:concept} illustrates this representation, while Table~\ref{tab:tokens} summarizes the five token types and their implementation in the current model.

\begin{table}[t]
  \centering
  \caption{Token types of \dlo{} and their representation in the current implementation. Each token encodes information across the observation window or prediction horizon, rather than representing a separate token for every observation time step. Contact tokens describe potential interactions, not only active contacts. The current model uses 155 tokens per cable.}
  \label{tab:tokens}
  \small
  \begin{tabular}{@{}p{2.3cm}p{5.4cm}p{7.0cm}@{}}
    \toprule
    Token type & Concept & Current implementation \\
    \midrule

    Cable node &
    Cable geometry and recent motion history, including position, velocity, and local frame &
    41 tokens per cable (one per node); each token encodes the position and velocity of its node over the 10 most recent time steps (\SI{0.2}{s}), resulting in 60 values per token \\

    Material &
    Physical cable properties, including stiffness, damping, mass, and rest shape &
    6 tokens per cable, one scalar each: bending stiffness, torsional stiffness, damping, density, length, and gripper height; constant along the cable \\

    Action and global &
    Planned robot motion, boundary conditions, and global scene information &
    25 action tokens (one per future time step, 7 values each) covering a prediction horizon of \SI{0.5}{s}, plus one global token for boundary conditions \\

    Rigid contact &
    Potential interactions with rigid surfaces, including geometry, normal, distance, relative velocity, and friction &
    41 tokens per cable (one per node), each describing the closest floor patch using 23 values, including geometry, distance history, relative velocity, friction, and multiscale proximity features \\

    Soft contact &
    Potential interactions with deformable objects, including cable--cable interactions and self-contact &
    41 tokens per cable (one per node), each describing the closest point on another cable using 30 values, including relative geometry, distance history, velocity, material properties, and multiscale proximity features \\

    \bottomrule
  \end{tabular}
\end{table}

Cable node tokens represent the recent motion of the DLO, while material tokens condition the prediction on its physical properties. 
These properties may, in principle, vary along the cable to represent heterogeneous materials and local rest configurations. 
Action tokens encode the planned manipulation, complemented by global tokens for boundary conditions.
Rigid- and soft-contact tokens extend this representation to interactions with the environment and other deformable objects, including self-contact. 
Unlike representations restricted to active contacts, they describe potential interaction partners that may come into contact during the prediction horizon. 

All token types are mapped into a shared embedding space and processed by a transformer. 
Through self-attention, the model can learn how material properties, robot actions, and contact interactions jointly influence cable motion without requiring explicit contact-force estimation.

\subsection{Forward prediction and inverse parameter identification}

The proposed representation supports both forward dynamics prediction and inverse parameter identification. 
Given recent cable states, physical properties, planned robot actions, and potential contacts, the model predicts cable motion over a specified horizon.

Conversely, unknown physical parameters can be identified from observed motion by minimizing the discrepancy between predicted and observed trajectories while keeping the model weights fixed:
\begin{equation}
  z^\ast = \arg\min_{z}
  \frac{1}{W}\sum_{w=1}^{W}
  \left\| f_\theta(x_w,a_w,z)-y_w \right\|^2 ,
  \label{eq:ident}
\end{equation}
where $z$ denotes the normalized parameter vector, $x_w$ the observed state history, $a_w$ the applied actions, and $y_w$ the observed future trajectory of window $w$. 
The predictor $f_\theta$ retains its trained weights $\theta$ throughout optimization.

The envisioned application combines both capabilities: identifying unknown DLO parameters from a few seconds of observed motion and subsequently using them to optimize a longer manipulation trajectory, without object-specific model retraining.

\section{Model Architecture and Implementation}
\label{sec:implementation}

This section presents the implementation of \dlo{}, including the design of the token representation, the transformer architecture, and the integration of rigid and soft contact interactions.

\subsection{Inputs and tokens}
\label{sec:impl_tokens}

Each prediction uses $K=10$ observation steps (\SI{0.2}{s}) and covers $H=25$ future steps (\SI{0.5}{s}) at \SI{50}{Hz}.
Longer forecasts are obtained through autoregressive rollouts, with contact tokens recomputed from the predicted states after each prediction.
Positions are expressed relative to the gripper at the last observed time step, and input features are normalized to comparable numerical ranges.

The cable is discretized into 41 nodes, each represented by a token containing its three-dimensional position and velocity over the 10 most recent time steps, resulting in 60 values per token.
Six material tokens encode bending stiffness, torsional stiffness, damping, density, cable length, and gripper height.
These parameters are normalized to $[-1,1]$, with logarithmic scaling applied to stiffness, damping, and density.
Robot actions are represented by 25 tokens containing seven values each, describing incremental and relative gripper motion.
The corresponding motion of the second cable's gripper is incorporated into the same action tokens.
A global token encodes whether the cable is fixed at both ends.

Rigid-contact tokens describe a floor patch beneath each cable node using 23 features, including surface geometry, distance history, relative velocity, friction, and proximity.
Similarly, soft-contact tokens describe the closest point on the interacting cable for each node of the predicted cable.
Each soft-contact token contains 30 features encoding the closest-point position, relative direction and distance history, relative velocity, proximity, and the interacting cable's material properties.

Both contact-token types represent potential rather than exclusively active contacts.
Contact proximity is represented by continuous distance-dependent features:
\begin{equation}
  \alpha(d,s)=\exp(-d/s),
  \label{eq:proximity}
\end{equation}
where $d$ denotes the distance between the potential interaction partners and $s$ determines the spatial range.
These features rescale the distance to help the model: they resolve the range close to contact and fade out for distant partners.
Several length scales cover both imminent and approaching contact, so that no explicit contact forces are required as inputs.

The resulting representation contains 155 tokens per cable, independent of the number of interaction partners.
Each cable node is assigned one rigid- and one soft-contact token, representing the closest potential contact point of each type.

\subsection{Network architecture}
\label{sec:impl_network}

\begin{figure}[t]
  \centering
  \includegraphics[width=\linewidth]{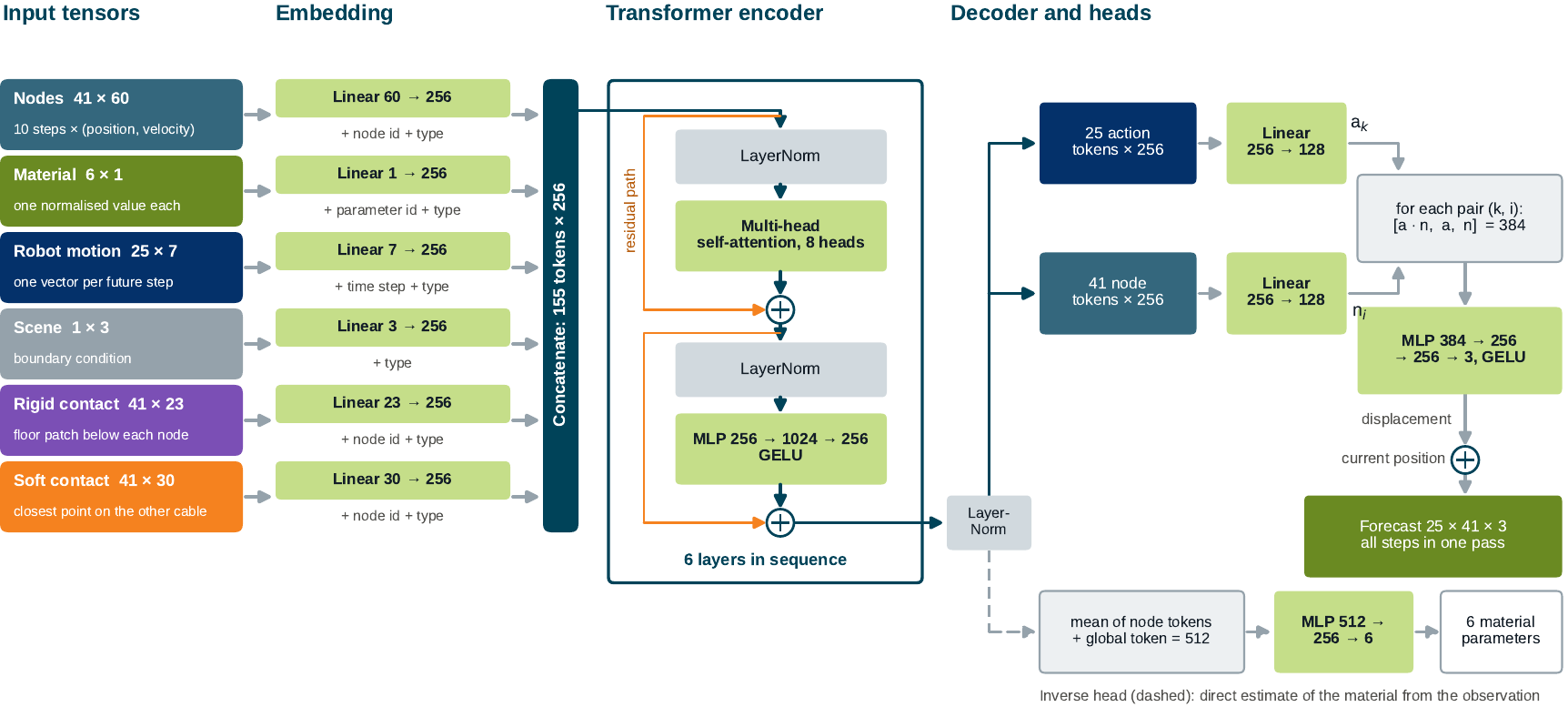}
  \caption{DLO-Anything network architecture. Token-specific embeddings are processed by a shared transformer encoder. A factorized decoder combines node and action representations to predict the complete cable trajectory over \SI{0.5}{s} in a single forward pass, while an additional head supports inverse parameter identification.}
  \label{fig:architecture}
\end{figure}

Figure~\ref{fig:architecture} illustrates the implemented architecture, comprising token-specific embeddings, a shared transformer encoder, a factorized motion decoder, and an inverse head for parameter identification.

Each token type is projected into a common embedding space of dimension $d=256$, with learned type and index embeddings preserving the distinction between token categories and their positions.
The resulting sequence is processed by six transformer encoder layers with eight attention heads and a feed-forward dimension of 1024.
Self-attention allows dependencies between cable nodes, material properties, robot actions, and potential contacts to be learned without prescribing fixed interaction structures.
This is particularly relevant for contact-rich dynamics, where the influence of different interaction partners changes throughout a manipulation.

The factorized decoder predicts the future motion of each cable node conditioned on the planned robot trajectory.
For this purpose, the transformer-encoded representation of each node is paired with that of each future action step, allowing the decoder to account for both node-specific dynamics and time-dependent manipulation effects.
Both representations are projected to 128 dimensions and combined through concatenation and element-wise multiplication.
A shared multilayer perceptron then predicts the displacement of node $i$ at future step $k$ relative to its last observed position:
\begin{equation}
  \hat{p}_i^{\,t+k}
  =
  p_i^{\,t}
  +
  \mathrm{MLP}(f_{k,i}),
  \qquad
  i=1,\ldots,N,\quad k=1,\ldots,H,
  \label{eq:decoder}
\end{equation}
where $f_{k,i}$ denotes the combined representation of node $i$ and future action step $k$.
By sharing the decoder across all node--time pairs, the architecture predicts the complete trajectory in a single forward pass without requiring separate prediction heads or autoregressive updates within the prediction horizon.

For inverse parameter identification, material tokens can be replaced by learned embeddings indicating unknown parameters.
An additional inverse head is designed to directly estimate the six physical parameters from the encoded node and global representations, potentially enabling faster parameter identification than the iterative optimization described in Eq.~\eqref{eq:ident}.
This allows the same encoder to support both dynamics prediction and parameter estimation.

The architecture contains approximately \num{5.2} million trainable parameters, with about \SI{91}{\percent} belonging to the transformer encoder.
The contact-token embeddings contribute less than \SI{1}{\percent} of the parameters but increase the number of attention pairs by a factor of approximately 4.5.
Table~\ref{tab:params} summarizes the parameter distribution.

\begin{table}[t]
  \centering
  \caption{Parameters of the default model (width 256, 6 layers, 8 attention heads).}
  \label{tab:params}
  \small
  \begin{tabular}{@{}llr@{}}
    \toprule
    Component & Architecture & Parameters \\
    \midrule
    Node, material, action and global embeddings &
    Linear projections and embeddings & \num{40192} \\
    Soft-contact embeddings &
    $30 \to 256$, partner action projection & \num{20736} \\
    Rigid-contact embeddings &
    $23 \to 256$ & \num{16896} \\
    Transformer encoder &
    6 layers, width 256 & \num{4739072} \\
    Factorized decoder &
    Node--action MLP & \num{230915} \\
    Inverse head &
    $512 \to 256 \to 6$ & \num{132870} \\
    \midrule
    Total & & \num{5180681} \\
    \bottomrule
  \end{tabular}
\end{table}

For the two-cable scenario, each cable is predicted by an instance of the same transformer architecture, conditioned on the other cable through soft-contact tokens.
The interaction partner's planned gripper motion is projected separately and added to the corresponding action tokens.
Both models predict their trajectories in parallel, using the same initial scene state.
This avoids iterative coupling between the models and explicit contact-force estimation, while allowing contact interactions to influence the predicted motion through the learned representation.

\section{Scalable training pipeline}
\label{sec:pipeline}

The simulation-based pipeline varies material properties, object geometry, robot trajectories, and contact conditions across episodes to generate training data for diverse dynamics and interactions.

\subsection{Simulation environment and parameter space}
\label{sec:pipeline_sim}

The data are generated with MuJoCo~\cite{mujoco}.
Each DLO comprises 40 capsule segments (41 nodes, radius \SI{3}{mm}) connected by bending and torsional stiffness.
A robot gripper moves one end horizontally at a fixed height above the floor; the point constraint transmits no torque.
A second object of the same construction hangs in a U-shape, fixed at both ends, across the path of the manipulated object.
The integration step is \SI{2}{ms}, states are recorded at \SI{50}{Hz}, and an episode lasts \SI{8}{s}, corresponding to 400 recorded steps.
Before recording, the robot remains at rest for \SI{3}{s} to allow the system to settle.

\begin{figure}[t]
  \centering
  \includegraphics[width=\linewidth]{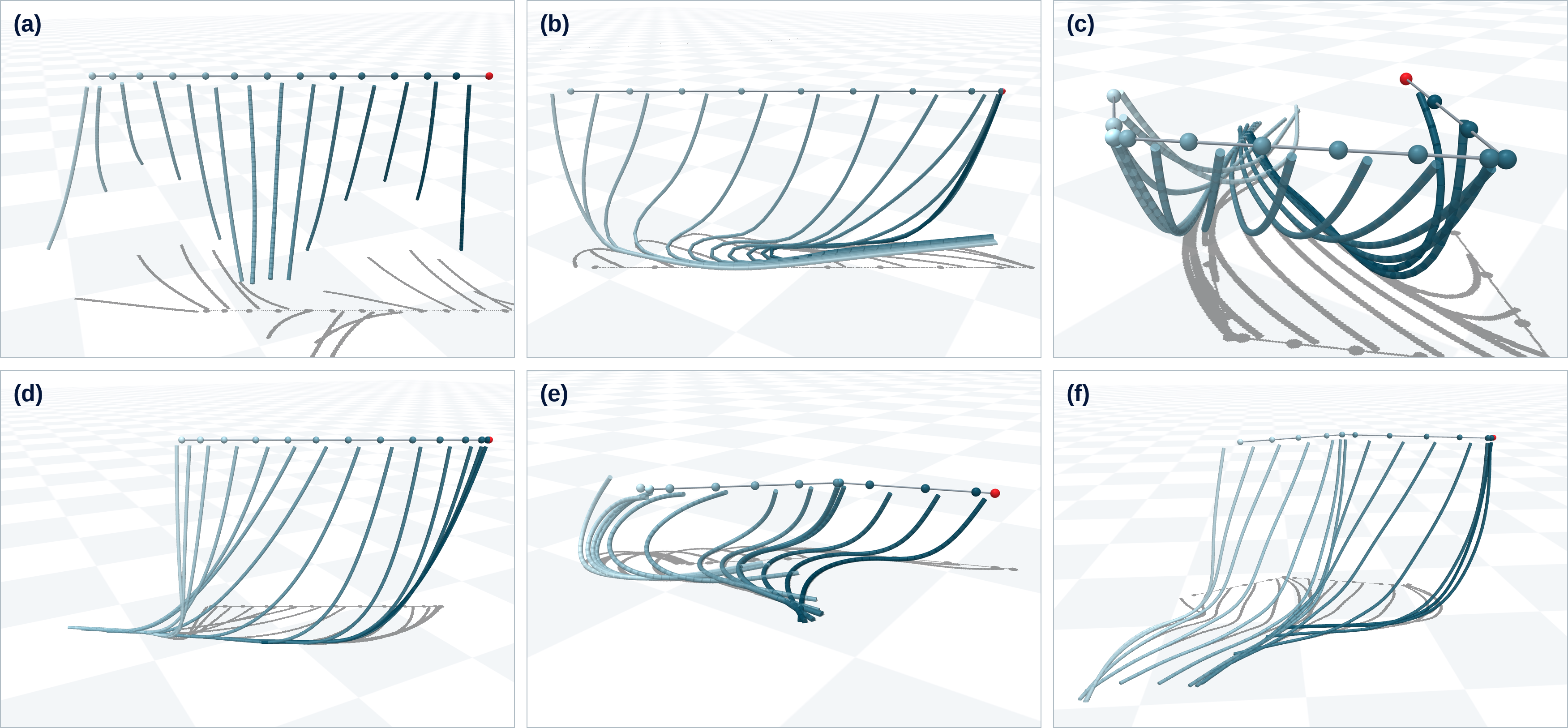}
  \caption{Six examples at the limits of the parameter ranges, each shown at 14 instants over \SI{2}{s} of motion (light: early, dark: late); the thin line is the path of the gripper. (a) Short, (b) long, (c) limp, (d) stiff, (e) light with a low gripper, (f) strongly damped.}
  \label{fig:cables}
\end{figure}

Four material parameters are sampled independently for each object and episode (Table~\ref{tab:ranges}).
Stiffness, damping, and density follow log-uniform distributions over one to three orders of magnitude; lengths and heights are constrained by the scene geometry to permit interaction.
Segment length is adjusted to represent objects between \SI{0.6}{m} and \SI{2.0}{m} with the same discretization.
Figure~\ref{fig:cables} illustrates the extremes of the parameter ranges.
Episodes are discarded if the simulation becomes unstable, a node moves more than \SI{10}{cm} between recorded steps, or the manipulated object winds more than once around the fixed one.

\begin{table}[t]
  \centering
  \caption{Simulation parameter ranges. Material properties are sampled independently for the two objects; lengths and heights are constrained by the scene geometry.}
  \label{tab:ranges}
  \small
  \begin{tabular}{@{}llll@{}}
    \toprule
    Parameter & Range & Unit & Sampling \\
    \midrule
    Bending stiffness   & \numrange{5e3}{5e6}    & \si{Pa}            & log-uniform \\
    Torsional stiffness & \numrange{1e3}{1e6}    & \si{Pa}            & log-uniform \\
    Damping             & \numrange{0.001}{0.08} & \si{N.m.s/rad}     & log-uniform \\
    Density             & \numrange{300}{4000}   & \si{kg/m^3}        & log-uniform \\
    Length, manipulated object        & \numrange{0.6}{2.0}   & \si{m} & scene-dependent \\
    Length, fixed object         & \numrange{0.8}{1.6}   & \si{m} & scene-dependent \\
    Gripper height              & \numrange{0.15}{1.6}  & \si{m} & scene-dependent \\
    Suspension height, fixed object & \numrange{0.3}{1.0} & \si{m} & scene-dependent \\
    \bottomrule
  \end{tabular}
\end{table}

\subsection{Training scenarios and datasets}
\label{sec:pipeline_data}

Robot trajectories are generated from eight motion categories to cover quasi-static dragging as well as faster swinging and whipping.
The first category comprises six designed patterns, including straight, angled, curved, and return motions.
The remaining seven categories comprise random waypoints, smooth splines, variable-speed curves, retraced paths, loops, weaving, and short bursts separated by pauses.
Within each category, path geometry, speeds of up to about \SI{1.2}{m/s}, and pauses are randomized across episodes.

\begin{figure}[t]
  \centering
  \includegraphics[width=\linewidth]{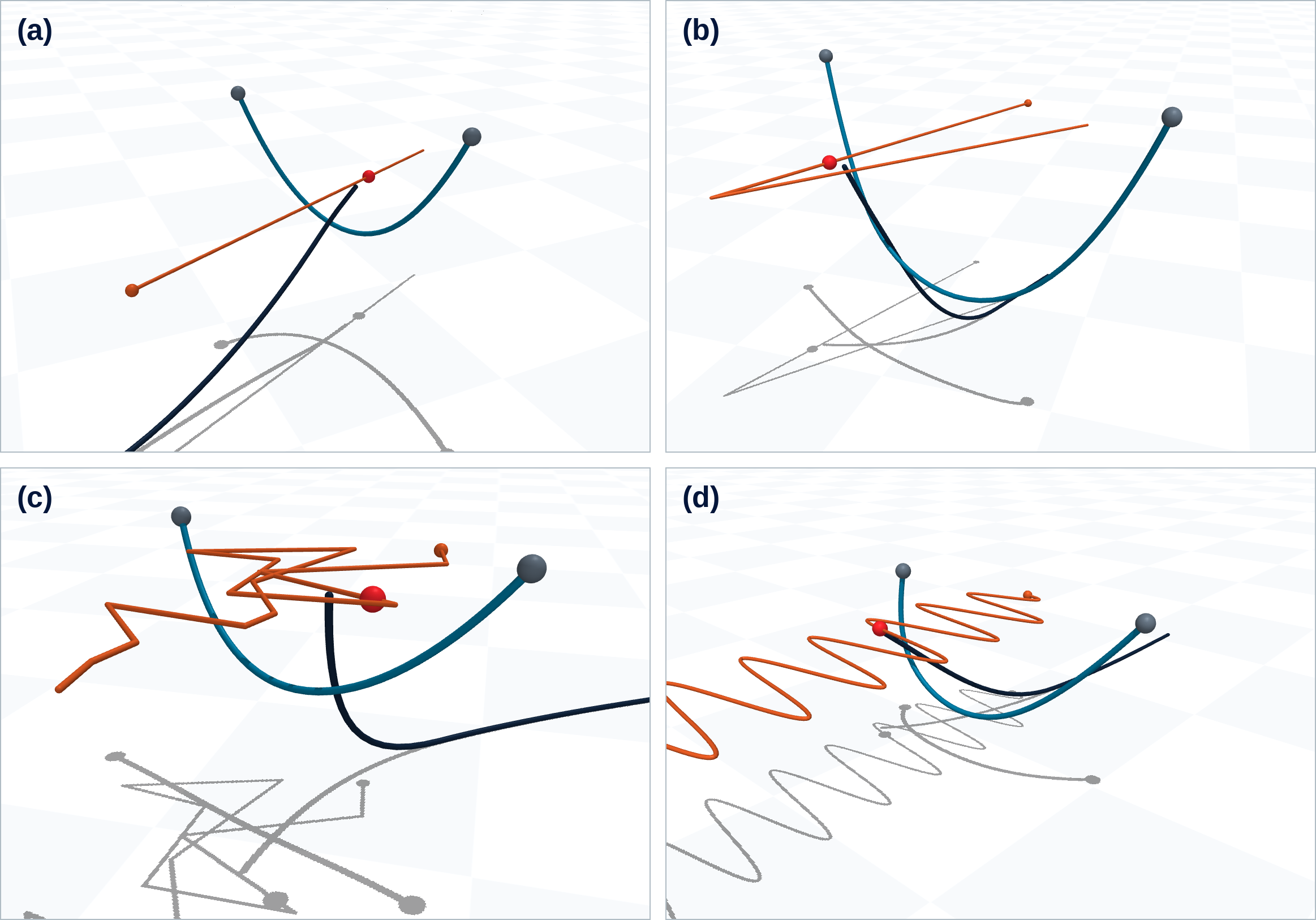}
  \caption{Robot trajectories in the two-object scenario. Each panel shows the manipulated object (dark) and the fixed object (teal, fixed points in gray) shortly after their first contact, together with the complete gripper path (orange). (a) Straight pass, (b) pass with a return, (c) bursts, and (d) a sinusoidal motion reserved for validation and testing.}
  \label{fig:motion_kinds}
\end{figure}

The contact scenario involves two interacting DLOs (Figure~\ref{fig:motion_kinds}).
The fixed object hangs in a U-shape with random orientation, while the gripper height is chosen to allow the manipulated object to cross it.
Material properties are sampled independently for both objects, and simulation contact stiffness is increased to limit interpenetration.
The six designed maneuvers comprise four single crossings (straight, angled, arc-shaped, or S-shaped), one crossing followed by a return, and one free swing without contact.

\begin{figure}[t]
  \centering
  \includegraphics[width=\linewidth]{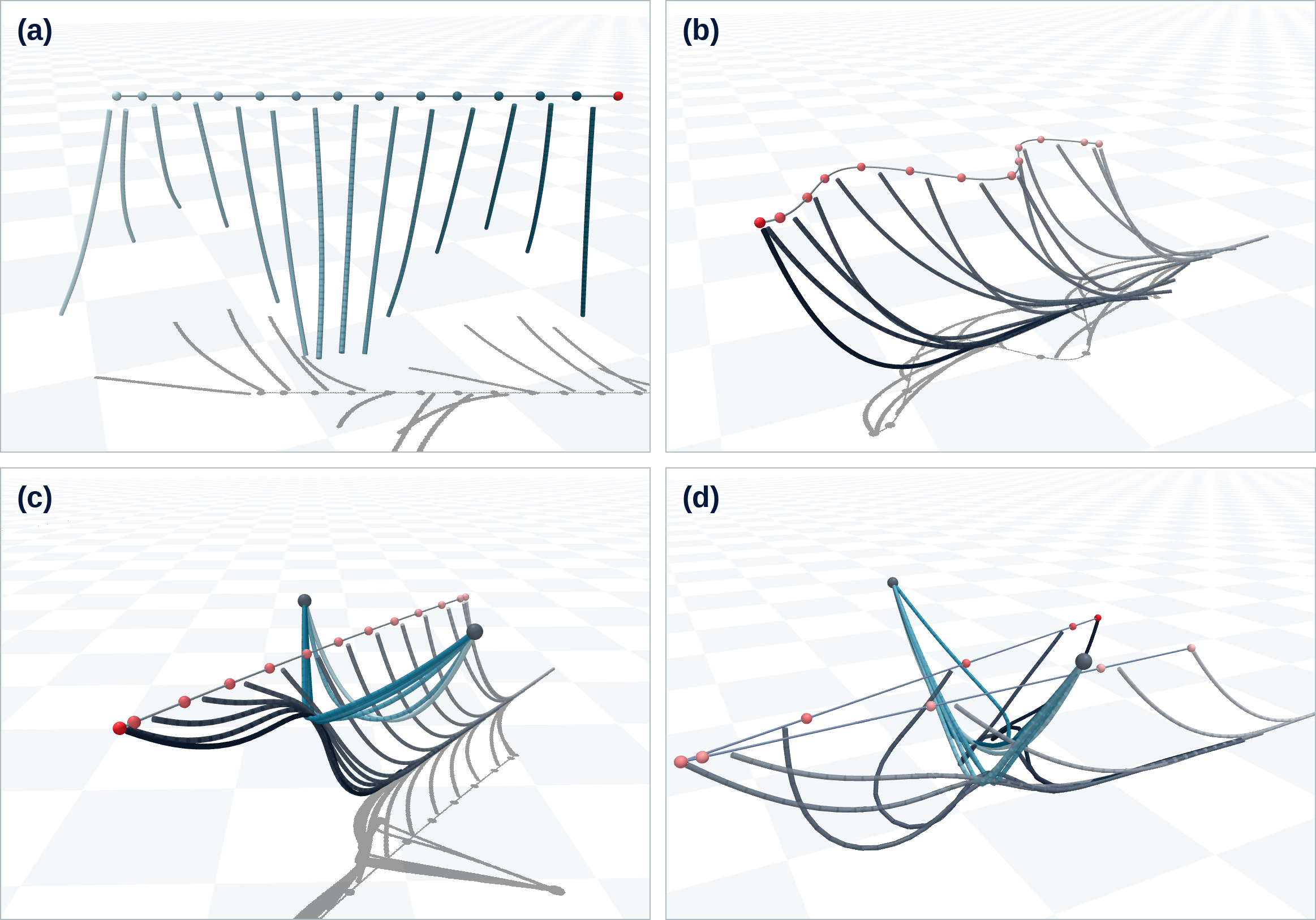}
  \caption{Representative episodes from four contact classes, shown at several instants (light: early, dark: late). The manipulated object is shown in dark blue, the fixed object in teal, and the robot path as a thin line. (a) Free motion, (b) floor contact only, (c) single crossing, and (d) reversal during contact.}
  \label{fig:cases}
\end{figure}

The resulting episodes are classified into four contact classes based on recorded forces and the maneuver (Figure~\ref{fig:cases}).
The \emph{free} and \emph{floor-only} classes involve no interaction between the two objects, whereas \emph{single-pass} and \emph{reversal} involve one or repeated crossings.
The latter are more challenging due to changing contact configurations and sliding directions.
Contact lasts approximately two to three seconds, with forces reaching \SI{0.4}{N} and displacements of the fixed object of about \SI{15}{cm}.
Reversal paths are approximately \SI{75}{\percent} longer and require the model to account for contact history across the change in direction.

To investigate the effects of training data volume and motion diversity, six dataset families (A--F) are constructed with different numbers of scenes and motion categories (Table~\ref{tab:data}).
Each included category contributes an equal share of scenes.
The largest datasets contain \num{24000} scenes, corresponding to 9.6 million recorded time steps or \SI{53}{h} of simulated motion for both objects.

Circular, sinusoidal, and jerky trajectories are excluded from training and reserved for validation and testing.
All datasets use disjoint random seeds, allowing each scene to be reproduced.

\begin{table}[t]
  \centering
  \caption{Training, validation, and test datasets. Each scene lasts \SI{8}{s} at \SI{50}{Hz} (400 time steps). Motion categories are equally represented within each training dataset.}
  \label{tab:data}
  \small
  \begin{tabular}{@{}llcl@{}}
    \toprule
    Set & Robot motions & Categories & Scenes \\
    \midrule
    A & designed patterns          & 1 & \numrange{371}{24000} \\
    B & A + random waypoints       & 2 & \numrange{371}{24000} \\
    C & B + spline curves          & 3 & \num{12000} \\
    D & C + varying speed          & 4 & \num{12000} \\
    E & D + retraced paths         & 5 & \num{12000} \\
    F & E + loops, weaving, bursts & 8 & \numrange{371}{24000} \\
    \midrule
    Validation & circle, sine, jerk & 3 & \num{303} \\
    Test       & circle, sine, jerk & 3 & \num{341} \\
    \bottomrule
  \end{tabular}
\end{table}

\subsection{Training procedure}
\label{sec:pipeline_training}

For each dataset in Table~\ref{tab:data}, a separate pair of models is trained using the same architecture, training procedure, and number of gradient steps.
Training windows are sampled at random positions within each scene, with half selected to contain contact to compensate for its limited duration.
The training objective is to minimize the mean squared error between predicted and simulated node positions across all nodes and future steps.

Training proceeds in two stages.
First, the two model instances are trained separately on individual prediction windows initialized from simulated states for approximately \num{30000} gradient steps (batch size 128, initial learning rate \num{3e-4}).
Both are initialized from a model pretrained on single-object motion, with additional embeddings for contact tokens.
Second, the models are trained jointly through \SI{6}{s} autoregressive rollouts comprising twelve consecutive prediction windows.
Each window starts from the preceding predictions, contact tokens are recomputed from both predicted states, and gradients propagate through the full rollout.
This stage uses \num{12000} gradient steps with a batch size of 24 rollouts and an initial learning rate of \num{1e-4}.
Both stages use AdamW with a weight decay of \num{0.01}, linear learning-rate warm-up followed by cosine decay, and gradient clipping at \num{1.0}.

Each training run was performed on a single NVIDIA RTX A6000 GPU with \SI{48}{GB} of memory, with both model instances trained in parallel during the first stage.
The median training time was approximately \SI{3.2}{h} for the first stage and \SI{3.7}{h} for the second, totaling about \SI{7}{h} per run, excluding initial pretraining.
These times were measured under shared GPU utilization, typically with two concurrent training runs.
Generating approximately \num{12000} simulation scenes required \SI{80}{min} using 36 CPU processes.

Prediction accuracy is evaluated on held-out robot motions using the mean Euclidean distance between predicted and simulated node positions, averaged over both objects, all nodes, and all steps of a \SI{6}{s} autoregressive rollout initialized from \SI{0.2}{s} of simulated context.

\section{Scaling Laws and Generalization}
\label{sec:scaling}

\begin{figure}[t]
  \centering
  \includegraphics[width=\linewidth]{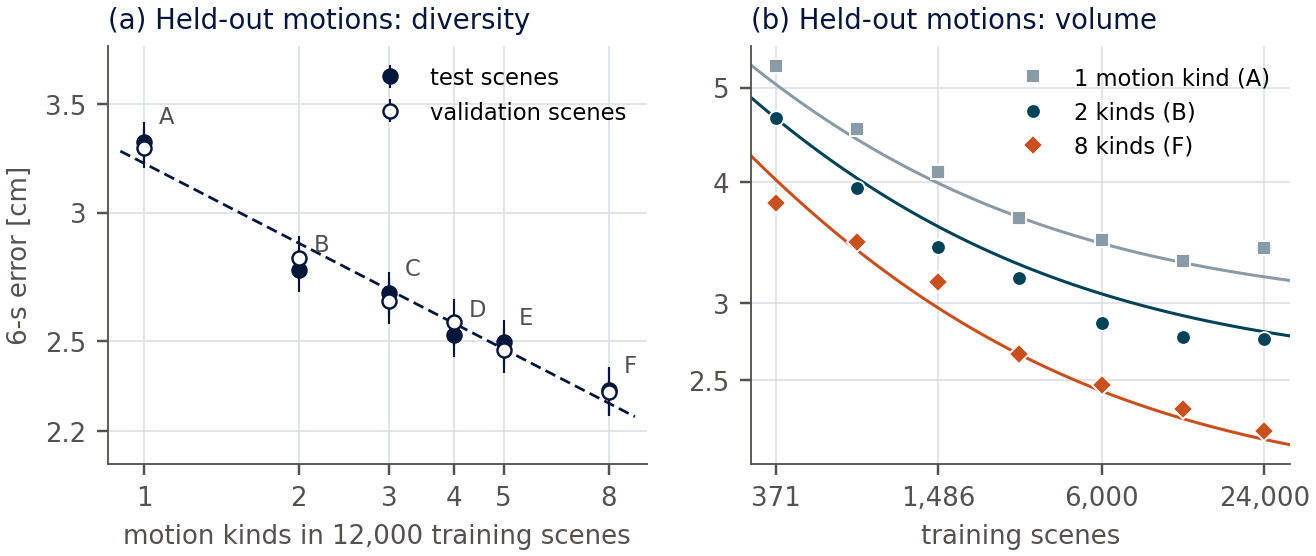}
  \caption{Scaling of mean prediction error over \SI{6}{s} autoregressive rollouts on unseen robot motions. (a) Effect of motion diversity at a fixed training set size of \num{12000} scenes, including a power-law fit. (b) Effect of training data volume for one, two, and eight motion categories, with curves representing Eq.~\eqref{eq:scaling}.}
  \label{fig:diversity}
\end{figure}

Figure~\ref{fig:diversity} shows how training data volume and robot motion diversity affect prediction accuracy on unseen manipulation trajectories.
At a fixed dataset size of \num{12000} scenes, increasing the number of motion categories from one to eight reduces the mean prediction error from \SI{3.32}{cm} to \SI{2.33}{cm}, an improvement of approximately \SI{30}{\percent} (Table~\ref{tab:ladder}).
The reduction follows an approximate power law, with each doubling of motion diversity reducing the error by about \SI{10}{\percent}.

\begin{table}[t]
  \centering
  \caption{Mean prediction error over \SI{6}{s} autoregressive rollouts on unseen robot motions, using \num{12000} training scenes with different numbers of motion categories.}
  \label{tab:ladder}
  \small
  \begin{tabular}{@{}lcccccc@{}}
    \toprule
    Set & A & B & C & D & E & F \\
    Motion categories & 1 & 2 & 3 & 4 & 5 & 8 \\
    \midrule
    Validation [cm] & 3.29 & 2.81 & 2.65 & 2.57 & 2.47 & 2.33 \\
    Test [cm]       & 3.32 & 2.77 & 2.68 & 2.52 & 2.50 & 2.33 \\
    \bottomrule
  \end{tabular}
\end{table}

Increasing the number of training scenes also improves prediction accuracy, but the gains diminish as the dataset grows (Figure~\ref{fig:diversity}b).
With only one motion category, the error decreases from \SI{5.26}{cm} at 371 scenes to \SI{3.32}{cm} at \num{12000} scenes, with no further improvement at \num{24000} scenes.
In contrast, eight motion categories achieve \SI{2.66}{cm} with only \num{2968} scenes, outperforming a dataset containing approximately eight times as many scenes but only one motion category.
This indicates that motion diversity is more important than data volume for generalization to unseen trajectories within the investigated setting.

Both effects can be described by a joint empirical scaling law:
\begin{equation}
  E(N,k) = a\,k^{-\beta}
  + c\,\Bigl(\frac{N}{1000}\Bigr)^{-\gamma},
  \label{eq:scaling}
\end{equation}
where $N$ denotes the number of training scenes and $k$ the number of motion categories.
The first term represents a diversity-dependent error floor, while the second describes the reduction in error with increasing data volume.
Fitting the model to 24 training runs yields $\beta=0.21$ and $\gamma=0.50$, with $R^2=0.98$.
The fitted relationship suggests that increasing motion diversity lowers the error floor that cannot be overcome by adding more scenes of the same motion categories.

The best model achieves a mean prediction error of \SI{2.22}{cm} across all nodes and time steps of a \SI{6}{s} autoregressive rollout on unseen robot motions.
Given the investigated DLO lengths of \SIrange{0.6}{2.0}{m}, this represents centimeter-level accuracy over a relatively long prediction horizon.
However, the error increases from \SI{0.71}{cm} after \SI{0.5}{s} to \SI{3.99}{cm} at the end of the forecast.
Contact interactions further increase prediction difficulty, with a mean error of \SI{2.54}{cm} in scenes involving contact between DLOs, compared with \SI{1.27}{cm} in contact-free scenes.

\section{Discussion}
\label{sec:discussion}

The results demonstrate a systematic relationship between training data diversity, data volume, and prediction accuracy on unseen robot motions.
In particular, increasing motion diversity lowers the error floor that cannot be overcome by adding more scenes of the same motion categories.
This suggests that the scalability of learned DLO dynamics models depends not only on the amount of training data but also on the diversity of physical interactions and manipulation scenarios represented during training.

However, the observed scaling relationship is limited to the investigated setting.
While additional motion categories consistently improved generalization, the underlying relationship between training and unseen trajectories remains unclear.
Improvements also varied across the held-out motions, indicating that the choice of training trajectories may matter beyond their number.
Moreover, diversity was varied only in robot motions within a single two-object scenario.
Whether similar scaling behavior extends to different scene configurations, contact conditions, and manipulation tasks remains to be investigated.

The empirical scaling law is subject to further limitations.
Each configuration was evaluated using a single training run with a fixed number of gradient steps, so the observed saturation with increasing data volume may partly reflect the training budget.
Furthermore, some additional motion categories resemble the held-out trajectories more closely than others, potentially contributing to the observed improvements.
Although the trend is also present with fewer categories, broader evaluations are needed to establish its robustness.

The current implementation also covers only part of the envisioned \dlo{} concept.
Interactions involving more than two DLOs, self-contact, branched structures, and spatially varying material properties have not yet been evaluated.
Furthermore, all experiments rely on simulated trajectories with exact state information and initially straight DLOs.
Real objects may exhibit permanent curvature and other properties not represented in the training data, which have been shown to affect prediction accuracy in practical handling scenarios.
Evaluating transfer to real-world manipulation and determining the amount of additional adaptation data required therefore remain important next steps.

\section{Conclusion and Outlook}
\label{sec:conclusion}

This paper introduced \dlo{}, an approach toward a foundation model for deformable linear objects, combining physics-conditioned dynamics prediction with a token-based representation of material properties, robot actions, and potential contacts.
The implementation demonstrates the prediction of interacting DLOs using two transformer models with \num{5.2} million parameters each, without explicit contact-force estimation.

The scaling study reveals that generalization to unseen robot motions depends strongly on training data diversity.
Increasing the number of motion categories from one to eight reduces prediction error by approximately \SI{30}{\percent} at a fixed dataset size, while additional data alone yield diminishing improvements.
A joint empirical scaling law captures the effects of both data volume and motion diversity, providing initial evidence for systematically improving DLO dynamics prediction through broader training data.

Future work will extend the training data to more diverse scene configurations, contact interactions, and DLO geometries, while investigating larger models and a shared architecture for multiple interacting objects.
A further priority is transferring the simulation-trained model to real-world manipulation scenarios and evaluating its use for trajectory optimization.

\section*{Code and Data Availability}

The source code, data generation pipeline, and evaluation results are available at \url{https://ludwigstr.github.io/DLO-Anything/}.

\bibliographystyle{unsrtnat}
\bibliography{references}

\end{document}